\documentclass[sigconf]{acmart}

\usepackage{array}     
\usepackage{tabularx}  
\usepackage{booktabs}  
\usepackage{soul}
\usepackage{url}
\usepackage{graphicx}
\usepackage{amsmath}
\usepackage{amsthm}
\usepackage{algorithm}
\usepackage{algorithmic}
\usepackage{enumitem}
\usepackage{multirow}
\usepackage{balance}
\usepackage{listings}
\usepackage{adjustbox}
\usepackage{xcolor}

\newcommand{\update}[1]{\textcolor{black}{#1}}

\newcommand{\name}{HuggingGraph}

\title{\name: Understanding the Supply Chain of LLM Ecosystem}

\author{Mohammad Shahedur Rahman}
\affiliation{
  \department{Department of Computer Science and Engineering}
  \institution{The University of Texas at Arlington}
  \city{Arlington}
  \state{TX}
  \country{USA}
}
\email{mxr7980@mavs.uta.edu}

\author{Peng Gao}
\affiliation{
  \department{Department of Computer Science}
  \institution{Virginia Tech}
  \city{Blacksburg}
  \state{VA}
  \country{USA}
}
\email{penggao@vt.edu}

\author{Yuede Ji}
\affiliation{
  \department{Department of Computer Science and Engineering}
  \institution{The University of Texas at Arlington}
  \city{Arlington}
  \state{TX}
  \country{USA}
}
\email{yuede.ji@uta.edu}

\copyrightyear{2025}
\acmYear{2025}
\setcopyright{cc}
\setcctype{by}
\acmConference[CIKM '25]{Proceedings of the 34th ACM International Conference on Information and Knowledge Management}{November 10--14, 2025}{Seoul, Republic of Korea}
\acmBooktitle{Proceedings of the 34th ACM International Conference on Information and Knowledge Management (CIKM '25), November 10--14, 2025, Seoul, Republic of Korea}
\acmDOI{10.1145/3746252.3761510}
\acmISBN{979-8-4007-2040-6/2025/11}

\begin{document}

\begin{abstract}
\noindent
{Large language models (LLMs)} leverage deep learning architectures to process and predict sequences of words, enabling them to perform a wide range of natural language processing tasks, such as translation, summarization, question answering, and content generation.
As existing LLMs are often built from base models or other pre-trained models and use external datasets, 
they can inevitably inherit vulnerabilities, biases, or malicious components that exist in previous models or datasets.
Therefore, it is critical to understand these components' origin and development process to detect potential risks, improve model fairness, and ensure compliance with regulatory frameworks. 
Motivated by that, this project aims to study such relationships between models and datasets, which are the central parts of the \textit{LLM supply chain}.
First, we design a methodology to systematically collect LLMs' supply chain information. 
Then, we design a new graph to model the relationships between models and datasets, which is a directed heterogeneous graph, having \update{\textit{402,654 nodes}} and \update{\textit{462,524 edges}}.
Lastly, we perform different types of analysis and make multiple interesting findings.


\end{abstract}






\maketitle
\section{Introduction}
\label{sec: introduction}

\noindent {Large language models (LLMs)} are AI models designed to understand and generate human language by learning patterns and relationships within extensive datasets~\cite{mirchandani2023large,shen2023slimpajama}, such as GPT (Generative Pre-trained Transformer)~\cite{yenduri2024gpt}, BERT (Bidirectional Encoder Representations from Transformers)~\cite{kenton2019bert}, and T5 (Text-To-Text Transfer Transformer)~\cite{bahani2023effectiveness}.
These models leverage deep learning architectures to process and predict sequences of words based on context, 
enabling them to perform a wide range of  tasks~\cite{bonner2023large}, such as, translation~\cite{lu2024llamax}, summarization~\cite{laban2023summedits}, question-answering~\cite{allemang2024increasing}, and content generation~\cite{agossah2023llm}.
LLMs usually have billions ore more parameters~\cite{merrick2024upscaling}, enabling them to generate high-quality text.


\begin{figure}[t]
  \centering
    \includegraphics[width=0.48\textwidth]{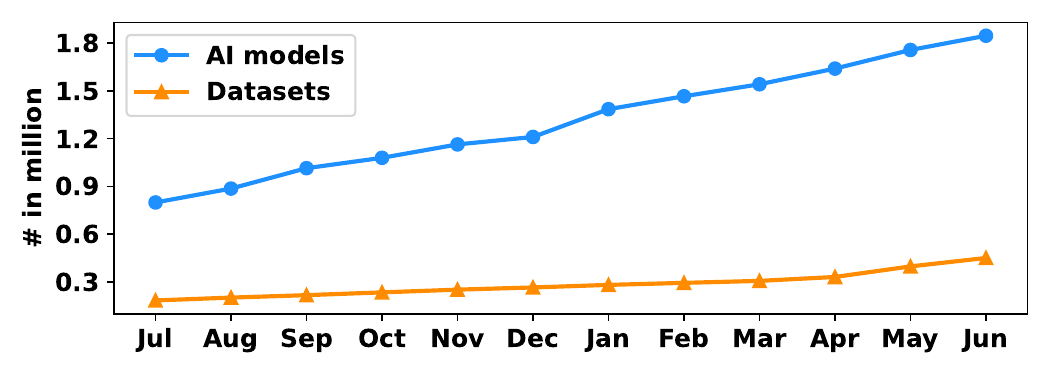} 
    \vspace{-6mm}
    \caption{The number of AI models and datasets (in million scale) on Hugging Face from July 2024 to June 2025.}
  \label{fig:model_increment}
\end{figure}
However, the increasing size and complexity of developing, training, and deploying cutting-edge LLMs demand extensive computational resources~\cite{xu2024survey} and large-scale datasets~\cite{villalobos2024will}.
This creates a significant barrier for researchers and practitioners, limiting their access to state-of-the-art models~\cite{mizrahi2024state}. 
As the demand for democratizing access to such LLM models continues to rise, platforms that host models and datasets have gained widespread popularity, e.g.,  Hugging Face~\cite{huggingface}, ONNX Model Zoo~\cite{onnx_model_zoo}, and PyTorch Hub~\cite{pytorchhub}.

Figure~\ref{fig:model_increment} shows the number of AI models and datasets (in million scale) on Hugging Face, one of the largest public AI model hosting platforms~\cite{huggingface}, from July 2024 to June 2025. By the end of June 2025, it has reached over 1.8M models and 450K datasets. In addition, the trend does not show any slowdown.
%
Such platforms provide user-friendly interfaces, APIs, and cloud-based infrastructures that enable researchers and developers to easily share, fine-tune, and deploy models without requiring extensive computational resources. 


Based on the tasks, these models can be classified into two broad categories, \textit{base models} and \textit{task-specific models}.
(i) \textit{Base models} are large, pre-trained models that can be fine-tuned for specific downstream tasks~\cite{tan2024challenges}. They are usually trained on vast datasets and are general-purpose, such as
GPT~\cite{yenduri2024gpt}, BERT~\cite{kenton2019bert}, and T5~\cite{bahani2023effectiveness}. 

(ii) \textit{Task-specific models} are modified versions of base models for a specific task. Taking Hugging Face as an example, there are four types of such models. First, \textit{fine-tuned models} adapt base models for specific tasks by training on additional task-specific datasets~\cite{zou2023comprehensive}. 
Second, \textit{adapter models} add lightweight and modular layers to the pre-trained models for specific tasks~\cite{hu2023llm}. 
Third, \textit{quantization models} trade off precision in numerical computations for accelerating inference and reducing memory consumption (e.g., using less precise model parameters)~\cite{velingkerclam}. 
Fourth, \textit{merged models} integrate multiple models into a single unified model by combining weights or configurations~\cite{akiba2025evolutionary}.
Besides, such platforms also host many \textit{datasets} used for training or adapting (e.g., fine-tuning) models~\cite{rottger2024safetyprompts}.

\subsection{Motivation}

\noindent As existing LLMs are often built from base models or other pre-trained models and use external datasets, 
they can inevitably inherit vulnerabilities, biases, or malicious components from previous models or datasets.
Thus, \textit{understanding these components' origin and provenance can help better detect potential risks, improve model fairness, and ensure compliance with regulatory frameworks}.

Motivated by that, this paper aims to study such relationships between models and datasets.
They are the central parts of the \textit{LLM supply chain}~\cite{wang2024large}, which refers to the entire lifecycle of developing, training, and deploying LLMs, similar to a traditional supply chain in manufacturing or software development~\cite{colakovic2021traditional,manufacturing2022review,chou2006software,sonatype2015software}. 
Such a supply chain can help to identify critical insights for both model evolution and dataset origin, as discussed below.

\textbf{Model evolution.} The study of the LLM supply chain gives a clear overview of how LLMs evolve from base models to fine-tuned variants, adapter integration, and quantization models. With that, one can easily keep track of them. For example, a use case is when a security vulnerability is found in one LLM, and we can quickly locate the potential models that might have the same 
vulnerabilities.

\textbf{Dataset origin.} 
This supply chain can also help to understand dataset origin used for training different models~\cite{shen2023slimpajama}. 
Dataset origin refers to the source from which the data is collected.
For example, for a fine-tuned model, we not only care about which dataset is used for fine-tuning but also what other datasets are involved in training the previous model.
Understanding such dataset origin helps to ensure that the dataset used is reliable, and  legally compliant.
\subsection{Contribution}

\noindent Our main contributions are threefold. 
First, we design a methodology to systematically collect the supply chain information of LLMs. In this paper, we mainly study the most popular AI platform, i.e., Hugging Face, but the same strategy applies to other platforms.
In particular, we use the APIs from the AI platform to collect the metadata about the hosted models and datasets.
To this end, we collected a large dataset as of \textit{June 30, 2025}. 



Second, with the collected metadata, we construct a new graph, named \textbf{\textit{LLM supply chain graph}}\footnote{The constructed graph: \url{https://github.com/SC-Lab-Go/HuggingGraph}}\footnote{A demonstration website: \url{https://ai-supply-chain.github.io/}}, to model the relationships between models and datasets. It is a directed heterogeneous graph, where a node denotes different types of models and datasets, 
and an edge denotes the dependency between them.
Together, this graph is able to accurately capture the LLM supply chain information.
To this end, we get a graph with \update{\textbf{402,654 nodes}} and \update{\textbf{462,524 edges}}. 



Lastly, we perform different types of analysis, including forward and backward analysis.
We study six research questions, including
(i) the properties of the LLM supply chain graph,
(ii) structural analysis,
(iii) supply chain relationships between AI models,
(iv) supply chain relationships between models and datasets,
(v) dynamic update evaluation, 
and (vi) generalizability to other AI platforms.

\textit{We hope this study can not only provide insights on LLM supply chain, but also raise awareness and future interests in this direction.}

\section{Preliminary}
\label{sec: background}

\noindent \textbf{LLM supply chain} encompasses the interconnected processes required for developing, deploying, and maintaining models~\cite{wang2024large}. This includes sourcing and preparing data to ensure high-quality and diverse datasets~\cite{wang2024large}. It also involves creating and training models~\cite{li2023large},
and making trained models available through APIs~\cite{singla2023empirical}. In addition, LLMs can undergo fine-tuning, adaptation, quantization, and merging processes in which they are tuned with domain-specific datasets to maximize performance on specific tasks~\cite{vm2024fine}, thus improving their accuracy and applicability.
This study mainly focuses on the relationships between models and datasets, which are the central parts of the whole LLM supply chain ecosystem.

\section{Methodology}
\label{sec: methodology}








\subsection{LLM Supply Chain Information Collection}
\label{sec:llm_sc_info_collect}

\noindent To analyze the LLM supply chain ecosystem, we need a large dataset with such information. 
Fortunately, platforms like Hugging Face provide some APIs that allow us to access the model and dataset and collect their metadata, which can be used to construct the LLM supply chain graph.

Table~\ref{tab:api_list} summarizes the four types of APIs we used.
In particular, (i) \textit{the model hub APIs} allow access to the hub of existing models, including searching and downloading the model and its metadata.
(ii) \textit{The dataset APIs} allow access to the datasets for discovery, metadata retrieval, and downloading.
(iii) \textit{The metrics APIs} allow access to the metrics for model evaluation, including metric discovery, metadata retrieval, and calculation.
(iv) One can use \textit{the search APIs} to search the name, tag, or other metadata for models and datasets.

\begin{table}[t]
\centering
\footnotesize
\caption{The APIs used to extract data from Hugging Face.}
\vspace{-2mm}
\setlength{\tabcolsep}{0.1cm}
\begin{tabular} {l|p{6.9cm}}
    \toprule
    \textbf{API Name} & \textbf{Description} \\   
    \midrule
    {Model hub} & {Access model hub to list, search, and download models and metadata.} \\ 

    \midrule
    Datasets & Access the datasets for discovery, metadata retrieval, and downloading. \\ 
    \midrule
    Metrics & Access metrics for model evaluation, e.g., metric discovery. \\
    \midrule
    Search & Search name, tag, or other metadata for model and dataset. \\
    \bottomrule
\end{tabular}
\vspace{-4mm}
\label{tab:api_list}
\end{table}


\textbf{Handling missing information.}
Accurate construction of the LLM supply chain graph depends on the quality of metadata from the LLM platforms (e.g., Hugging Face), which might suffer from missing or incomplete data.
To address it, we design two techniques, i.e., cross-reference links, and textual pattern extraction.

\textit{(i) Cross-reference links.} 
The model or dataset description could miss the supply chain data fields for API queries, which could be embedded within statically or dynamically rendered HTML pages. In this example URL \url{https://huggingface.co/models?other=base_model:finetune:meta-llama/Meta-Llama-3-8B}, one can tell the model ``\textit{Meta-Llama-3-8B}'' is fine-tuned from the model in the previous webpage.
To capture such information, we cross-reference the links of the filtered model listing webpages, extract model identifiers, enable the reconstruction of supply chain graph edges, and recursively trace model lineage from the leaf node. 
This scraping step complements API-based extraction and is only employed when reverse dependency data is otherwise inaccessible.

\textit{(ii) Textual pattern extraction.} 
When structured metadata is absent, the model and dataset cards might mention dependencies in unstructured text descriptions. To capture that, we employ a named entity recognition (NER) method~\cite{vasiliev2020natural, li2020survey}
to extract the dependency relationships from the text. For example, the textual phrase like ``fine-tuned from Llama-2'' contains the \textit{fine-tuned} keyword, which implies from which model this model is actually fine-tuned. Similarly, we also look into other words, such as, ``train'', and ``adapt''.

\subsection{LLM Supply Chain Graph}
\label{subsec:llm_SC_Grapg}


\noindent The collected metadata can be accurately modeled by the graph data structure.
It is a directed heterogeneous graph,
where a node denotes different types of datasets and models, including base, fine-tune, adapter, quantization, and merge models. An edge denotes the dependency relationship between them, including model-model, dataset-dataset, and model-dataset relationships.

\begin{figure}[t]
  \centering
    \includegraphics[width=0.48\textwidth]{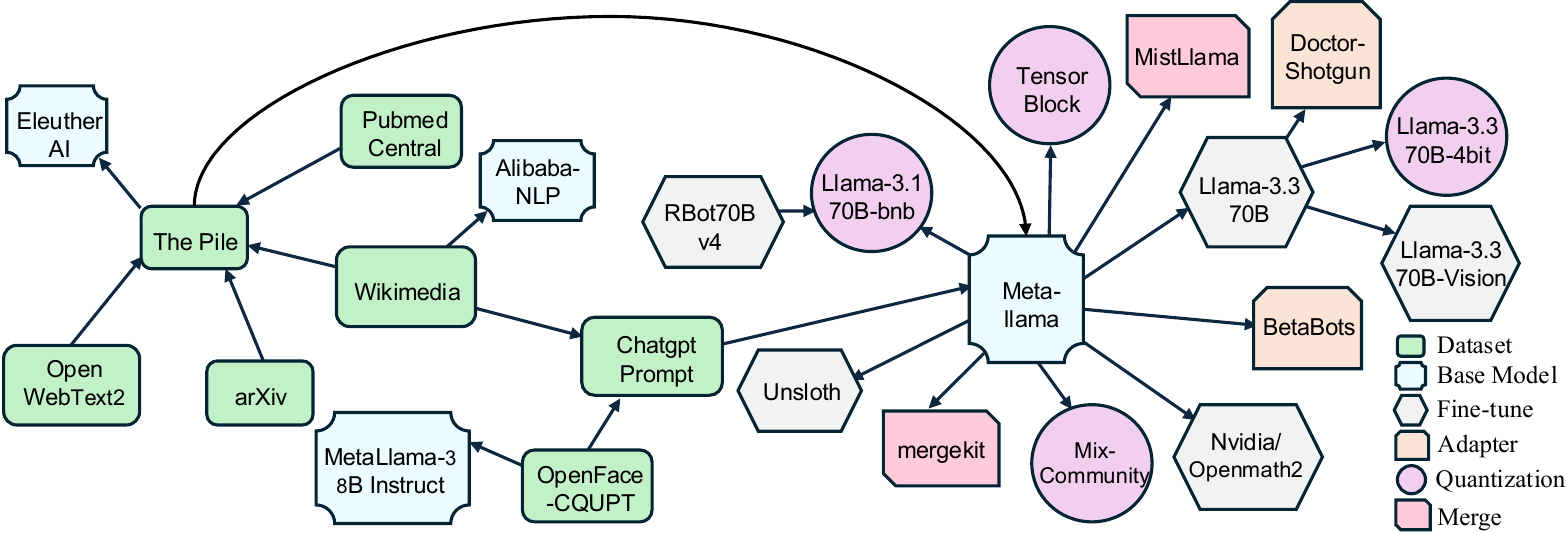}
    \caption{An example subgraph centering on base model ``\texttt{Meta-llama}'' from the complete LLM supply chain graph.
    }
  \label{fig:sample_subgraph}
\end{figure}

Figure~\ref{fig:sample_subgraph} shows a simplified supply chain subgraph centering on a base model \textit{Meta-llama}. 
\textit{(i) Model-model relationship.} To further identify the supply chain relationship, we will check the relevant data fields. 
In particular, given a model in Hugging Face, there are data fields ``\texttt{finetune}'' that show which models are fine-tuned from this model.
Similarly, ``\textit{adapter}'', ``\textit{quantization}'', \update{and ``\textit{merge}''} show which models are adapted, quantized \update{or merged} from them, respectively.
With such information, we can construct the supply chain relationship between the models.
As shown in Figure~\ref{fig:sample_subgraph}, model \textit{Llama-3.3-70B} is fine-tuned from the base model \textit{Meta-llama}. Then, it is used by the models \textit{Doctor-Shotgun} and \textit{Llama-3.3-70B-4bit} to generate an adapter and quantization model, respectively.

\textit{(ii) Dataset–dataset relationship.}  
The datasets within the LLM supply chain might overlap, build upon, or extend from each other. For example, a dataset may be a subset or modified version of another. 
To capture such information, we connect them with two types of edges.
\textit{(1) Subset relationships} arise when a dataset is explicitly documented as a subsample or partition of another. For instance, a dataset named \textit{``C4\_200M''} 
is described as a subset of \textit{``C4''}.  
\textit{(2) Modified versions} represent updates or enhanced variants of existing datasets. For example, \textit{``TruthfulQA\_v2''} incorporates corrections and improvements over an earlier version, \textit{``TruthfulQA\_v1''}.  

\textit{(iii) Model-dataset relationship.} 
To capture this,
we use the metadata from both models and datasets. The metadata of a model might specify the datasets used for training or adapting. However, not all the models disclose such information. 
Fortunately, we find the metadata of a dataset contains a field of ``trained\_fine\_tune\_models'' on this dataset. Thanks to that, we can capture the accurate model and dataset relationship. 
In Figure~\ref{fig:sample_subgraph}, the datasets \textit{The Pile} and \textit{Chatgpt-prompt} have directed edges to model \textit{Meta-Llama}, meaning that both datasets are used to train the model.

\subsection{Supply Chain Graph Analysis} 

\noindent
This supply chain graph can help to understand the transformational processes of the models and datasets. 
In particular, we can understand how base models evolve into their variants, including fine-tuned, adapted, quantized, or merged models, and vice versa. Similar observations can be made for datasets. This would provide a clear view of how the base model (or dataset) is transformed for performing a particular task. In particular, we mainly perform two types of analysis, i.e., forward and backward analysis.

\textbf{Forward analysis} is the method of traversing the supply chain graph following 
the dependency edges of a chosen node in a forward-going way. This node (known as the root/source node) can be a dataset, a base model, a fine-tuned model, an adapter, a quantized, or a merged model. 
In particular, given a source node, we apply the graph traversal algorithm (e.g., breadth-first search (BFS)~\cite{Knuth1974}) to traverse all the nodes (including both models and datasets) in a level-by-level pattern. 
To that end, this forward analysis will identify all the nodes that are reachable from the source node.

\textit{Model analysis example.}
In Figure~\ref{fig:sample_subgraph}, we analyze the forward supply chain of \textit{Meta-Llama}.
In particular, we identify \update{four} distinct forward paths:
(i) Base model (\textit{Meta-llama}) $\rightarrow$ fine-tuned model (\textit{Llama-3.3-70B})  $\rightarrow$ another fine-tuned model (\textit{Llama3.3-70B-Vision}).
(ii) Base model (\textit{Meta-llama}) $\rightarrow$ adapted model (\textit{Doctor-Shotgun}).
(iii) Base model (\textit{Meta-llama}) $\rightarrow$ fine-tuned model (\textit{Llama-3.3-70B})  $\rightarrow$ quantization model (\textit{Llama-3.3-70B-4bit}).
\update{(iv) Base model (\textit{Meta-llama}))  $\rightarrow$ merged model (\textit{MistLlama}).}
These paths show the evolution trajectory of the base model \textit{Meta-Llama}, showcasing its progressive specialization and adaptation for various tasks.

\textit{Dataset analysis example.}
For the dataset, our supply chain analysis shows how different datasets connect and form a new dataset.
This combination creates flexible resources that show how models perform in various areas.
In Figure~\ref{fig:sample_subgraph}, the dataset \textit{The Pile} is composed of multiple subsets, including \textit{Wikimedia}, \textit{arXiv}, \textit{OpenWebtext2}, and \textit{Pubmed Central}. Together, these datasets form a unified corpus that serves as training data for models like \textit{Meta-LlaMA}.

\textbf{Backward analysis} is the method of traversing the supply chain graph following the edges in a backward way. 
We accomplish this by traversing the graph also with BFS~\cite{Knuth1974}, starting from the selected node and following the incoming edges. To that end, this backward analysis will identify all the nodes that can reach the source node.

\textit{Model and dataset analysis example.}
In Figure~\ref{fig:sample_subgraph}, analyzing the backward supply chain of model \textit{RBot70Bv4}, we trace its lineage through its development stages. This model is fine-tuned from \textit{Unsloth}, which in turn originates from its base model, \textit{Meta-Llama}, and the datasets used to train the base model are \textit{The Pile} and \textit{Chatgpt-prompts}. 
Through this analysis, we establish the backward path, starting from target model, \textit{RBot70Bv4}, and tracing to its base model, \textit{Meta-Llama}, revealing dependencies and transformations in its development.
Similar methods can be used for dataset analysis.


\subsection{Accommodating Dynamic Update} 
\label{sec:temporal_updates}

\begin{figure}[t]
  \centering
    \includegraphics[width=0.48\textwidth]{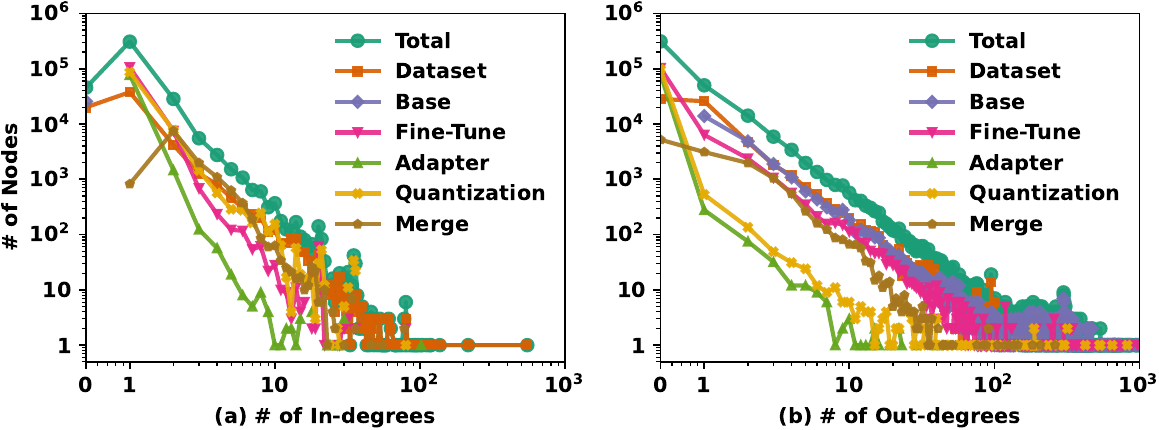} 
    \vspace{-7mm}
    \caption{
      (a) Indegree and (b) outdegree distribution. X-axis shows degrees, and Y-axis shows node count (log scale).
      } 
  \label{fig:degree_distribution}
  \vspace{-4mm}
\end{figure}

\noindent 
The hosted models and datasets on AI platforms are growing fast as new models are being developed every day. 
For example, between June 25 and July 15, 2025, we observed 80,703 new models (approximately 3,843 per day), 27,405 new datasets (approximately 1,305 per day) on Hugging Face, which is just one of the many AI platforms.
Therefore, we need to accommodate the dynamic update to accurately manage and analyze the AI supply chain. 



Particularly, {\name} accommodates the dynamic updates in three steps.
\textit{(i) Scoping updated models or datasets.} At a time $t$, we keep a copy of the hosted models and datasets with their IDs. 
When it evolved to $t+1$, we get another copy of the hosted models and datasets with their IDs.
The difference between them shows the updated models and datasets, including newly added or deleted.
In our current implementation, we are keeping the update on a daily basis.
\textit{(ii) Metadata collection for the updated models or datasets.}
For the identified updated models or datasets, we will collect their metadata using the same strategy as discussed in Section~\ref{sec:llm_sc_info_collect}.
To this end, we get the updated dependencies between models and datasets. 
That is, for the update at time $t+1$ compared to $t$, it can be represented as $\Delta_{t+1}$.
\textit{(iii) $\Delta$-based dynamic graph update.}  
Given the newly updated dependency $\Delta_{t+1}$, and let $G_t$, $G_{t+1}$ denote the graph at time $t$, $t+1$, respectively, then $G_{t+1}=G_{t}\cup \Delta_{t+1}$.

\section{Experiment and Finding}


\noindent To deeply understand the relationships between models and datasets, we study six critical research questions (RQs) as below.


\begin{itemize}[leftmargin=*]
    \item \textbf{RQ \#1}: What are the properties of LLM supply chain graph?
    \item \textbf{RQ \#2}: What structural patterns emerge?
    \item \textbf{RQ \#3}: What are the supply chain between LLM models? 
    \item \textbf{RQ \#4}: What are the relationships between models and datasets?
     \item \textbf{RQ \#5}: What insights can be gained from the dynamic updates?
    \item \textbf{RQ \#6}: How can {\name} be applied to other platforms?
        \vspace{-2mm}

\end{itemize}

\subsection{RQ \#1: Supply Chain Graph Properties}
\label{sec:sc_graph_properties}

This research question aims to understand the critical properties of LLM supply chain graph, i.e., graph basics and degree distribution.

\textbf{Graph basics.}
The collected supply chain graph is a \update{medium-scale} directed heterogeneous graph with \textbf{402,654 nodes} and \textbf{462,524 edges} 
as of \textit{June 30th, 2025}.
In particular, there are six different types of nodes, including 28,384 base models, 115,211 fine-tuned, 79,254 adapters, 98,143 quantization models, 13,028 merges, and 68,634 datasets. The average degree is about 1.15, denoting that it is a very sparse graph. 
\update{Furthermore, we identified substantial metadata missing. As of June 30, 2025, among 1.8 million models, only 50,156 (2.79\%) provides a model tree, while $\sim$550K models lack any metadata beyond their names, and nearly another 400K models are empty. Similarly, of the 450K datasets, only 68,634 (15.26\%) provide dataset cards, leaving $\sim$380K datasets without any metadata.}

\textit{This highlights a broader issue in the AI community, where a significant number of models and datasets lack consistent and structured documentation on the supply chain. This reflects the need of more transparent disclosure.}

\textbf{Degree distribution}
shows how node degrees (the number of edges connected to a node) are distributed across the graph. 
Figure~\ref{fig:degree_distribution} illustrates the indegree and outdegree distribution of the graph. We show not only the total distribution but also the distribution of six types of nodes, including base models, fine-tuned models, adapter models, quantization models, merged models, and datasets. 

We observe that the degree distribution in our supply chain graph is \textbf{heavy-tailed}.
In particular, the indegree distribution shows a large spread across different categories. The outdegree distribution follows a similar pattern but may differ in specific cases (e.g., adapters seem to have a more restricted degree distribution). 
This
{heavy-tailed behavior} suggests that most nodes have low degrees, while a few central nodes (hubs) dominate the graph. 
In particular, the dataset \textit{macrocosm-os/images} has the highest indegree 550, and dataset \textit{Mistral-v0.1} from ``mistral AI'' has the highest outdegree 1,093.
Specifically, the base models act as high-degree hub nodes as they are heavily used by other task-specific models.




\vspace{1mm}
\noindent\textbf{Finding \#1}: \textit{The LLM supply chain graph is medium-scale, sparse, and heavy-tailed distribution. However, a significant number of models and datasets lack metadata, highlighting the need for more transparent supply chain documentation.}

\subsection{RQ \#2: Supply Chain Structural Analysis}
\label{sec:structural_analysis}

\begin{figure}[t]
  \centering
    \includegraphics[width=0.48\textwidth]{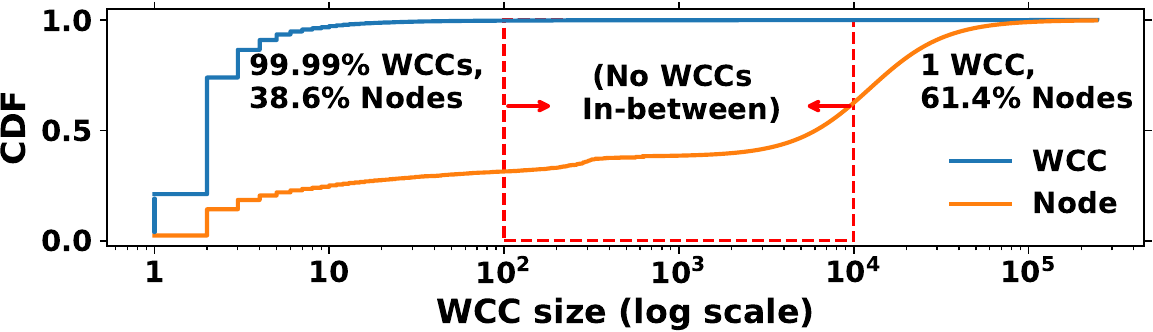} 
    \vspace{-7mm}
    \caption{Cumulative distribution function (CDF) of WCC.
    }
    \vspace{-5mm}
  \label{fig:wcc_analysis}
\end{figure}



\noindent{This research question aims to understand the topology and evolution of the LLM supply chain. To achieve that, we analyze the structural properties with connectivity and community analysis.}


\textbf{Connectivity analysis}.
\update{We computed weakly connected components (WCCs), which identify a maximal subset of nodes that remain connected when edge directions are ignored~\cite{ji2020aquila,ji2020swarmgraph}. The total number of WCCs in our supply chain graph is 44,908.}

Figure~\ref{fig:wcc_analysis} shows the cumulative distribution function (CDF) of the WCC distribution. 
We observe
(i) the largest WCC covers 247,244 nodes, accounting for 61.4\% of all the nodes. It reflects the dense interconnections that pervade the ecosystem. This vital element is essential for effective information sharing, resource allocation, and structural support, and is the base of the ecosystem. 
In the largest WCC, major models are included, such as \textit{Gemma-2B}, \textit{DistilBERT}, and \textit{GPT-2}.
(ii) In contrast, the remaining WCCs collectively hold 38.6\% of the nodes,
with most having 1, 2, or 3 nodes. This indicates a fragmented outer edge characterized by specialized models, rare datasets, or active experimental projects. The prevalence of these small, isolated pieces suggest niche attempts that lack integration with the overall system.
For example, \textit{zhongqy/RMCBench} (benchmarking dataset) and \textit{yigit69/bert-base-uncased-finetuned-rte-run\_3} (recognizing textual entailment task) remain disconnected due to limited reuse or insufficient metadata.

\update{We also computed strongly connected component (SCC)~\cite{ji2018ispan}, a maximal subgraph in which every node is reachable from every other via directed edges, identifying 398,198 SCCs. Remarkably, only 591 of these are non‑trivial (size > 1), collectively encompassing 5,047 nodes (1.25\% of the graph).
The largest SCC comprises 478 dataset nodes, among them \textit{tree‑of‑knowledge} and \textit{OpenHermes‑2.5}, forming a tightly‑knit cluster. By contrast, the remaining 99.46\% of nodes each reside in trivial (size‑1) SCCs.}

\textbf{Community detection} identifies node groups with dense internal connections. In LLM supply chains, it reveals aligned subgraphs reflecting reuse patterns and task-specific assets.
We apply the Louvain method~\cite{de2011generalized}, a greedy algorithm that maximizes modularity to find densely connected communities. 
Higher modularity indicates stronger intra-community connectivity.


Table~\ref{tab:communities} summarizes the top-10 communities.
(i) 
We find that each of the top-10 communities achieve a high modularity score of 0.96, indicating strong intra-community connectivity. Collectively, these communities span a wide range of functional domains, underscoring the presence of well-defined, task-aligned clusters within the ecosystem.
(ii) The largest community consists of 9,390 nodes and attains a modularity score of 0.96, indicating an extremely cohesive internal structure. It revolves around base models like \textit{OLMoE} and \textit{CausalLM}, and general-purpose datasets like \textit{prompt-perfect}. This suggests a densely connected cluster facilitating widespread reuse and fine-tuning.
(iii) Several other communities reflect clear task-based segmentation. For instance, community 2 (7,388 nodes) focuses on solving mathematical problems, with models like \textit{qwen2.5\_math}  and datasets like \textit{Marco-o1}. Similarly, community 6 (4,554 nodes) centers on instruction-tuning, with model \textit{MedLlama-3-8B} and dataset \textit{dpo-mix}.

\begin{table}[t]
\centering
\footnotesize
\caption{Top-10 Louvain communities sorted by size.}
\vspace{-4mm}
\begin{tabular}{c r p{2.3cm} p{2.3cm} c}
\toprule
\textbf{ID} & \textbf{Size} & \textbf{E.g. models} & \textbf{E.g. datasets} & \textbf{Modularity} \\
\midrule
1  & 9,390  & OLMoE, CausalLM      & prompt-perfect    & 0.96 \\
2  & 7,388  & qwen2.5\_math        & Marco-o1          & 0.96 \\
3  & 6,989  & Wanxiang             & smartllama3.1     & 0.96 \\
4  & 6,813  & tinyllama            & Llama-1B          & 0.96 \\
5  & 5,163  & Qwen2.5-32B          & Matter-0.2        & 0.96 \\
6  & 4,554  & MedLlama-3-8B        & dpo-mix           & 0.96 \\
7  & 4,262  & Electra, ArliAI      & MixEval           & 0.96 \\
8  & 3,947  & aesqwen1.5b          & llava             & 0.96 \\
9  & 3,829  & bert                 & TORGO             & 0.96 \\
10 & 3,828  & Mistral              & vicuna\_format    & 0.96 \\
\bottomrule
\end{tabular}
\vspace{-2mm}
\label{tab:communities}
\end{table}

 \vspace{1mm}
\noindent\textbf{Finding \#2:}
\textit{The LLM supply chain graph features a dominant core (61.4\% of nodes), while high modularity (0.96) reveals task-aligned, semantically coherent communities amid a fragmented periphery.}

%
\subsection{RQ \#3: Supply Chain Analysis of Models}



\noindent This research question aims to provide a holistic view of the dependencies between the models within the LLM supply chain, particularly from both base and task-specific models.


\textbf{Base model impact.}
We would like to understand the impact of base models. Here, we quantify the impact of a base model as the number of task-specific models that depend on it. The more dependencies, the larger the impact it has. We start with a base model and perform forward analysis by computing BFS following the outgoing edges.
This leads to a forward subgraph, which denotes all the models that depend on the base model, including fine-tuned, adapted, quantized, or merged models.

Table~\ref{tab:forward_model} shows the top-10 base models sorted by the forward subgraph size, which is the number of impacted task-specific models.
We find (i) a base model can significantly impact the LLM supply chain ecosystem. For example, \textit{Llama-3.1-8B} is a base model from Meta used for efficient text generation, code assistance, and research~\cite{he2024llama}. Due to its relatively small size, which allows for deployment in resource-constrained environments, making advanced AI accessible to broader stakeholders~\cite{grattafiori2024llama}. It has generated up to 7,544 models, including 1,710 fine-tuned versions, 1,542 adapters, 3,473 quantizations, and 1,693 merged models tailored to specific tasks.
(ii) For fine-tuning, the base model \textit{Mistral-7B-v0.1} has been fine-tuned the most, totaling 2,105. 
It is a faster, lighter \textit{Mistral} model trained by Mistral AI with grouped-query and sliding-window attention, enabling efficient text generation, NLP, and code assistance on consumer hardware for low-latency tasks~\cite{thakkar2023comprehensive}.

\begin{table}[t]
\centering
\footnotesize
\setlength{\tabcolsep}{0.12cm} 
\caption{Top-10 base models sorted by forward subgraph size.}
\vspace{-4mm}
\begin{tabular}{lrrrrrr}
\toprule
\textbf{Base model} & 
\textbf{Total} & 
\textbf{\begin{tabular}[c]{@{}r@{}}Fine-\\ tune\end{tabular}} & 
\textbf{Adapter} & 
\textbf{\begin{tabular}[c]{@{}r@{}}Quanti-\\ zation\end{tabular}} & 
\textbf{Merge} &
\textbf{Level} \\
\midrule
Llama-3.1-8B             & 7,544 & 1,710 & 1,542 & 3,473 & 1,693 & 25 \\
Mistral-7B-v0.1          & 6,744 & 2,105 & 2,187 & 1,435 & 1,254 & 27 \\
Qwen2.5-7B               & 6,733 & 1,972 & 1,764 & 2,516 & 1,132 & 11 \\
Meta-Llama-3-8B          & 5,633 &  967 & 1,511 & 2,220 & 1,967 & 21 \\
Llama-3.1-70B            & 4,063 &  698 &  281 & 2,075 & 2,519 & 11 \\
Qwen2.5-32B              & 3,909 & 1,086 &  158 & 2,311 & 1,049 & 12 \\
Qwen2.5-1.5B             & 3,645 & 1,300 & 1,290 &  949 &  248 &  8 \\
Qwen2.5-0.5B             & 3,521 & 1,669 & 1,006 &  810 &   46 & 11 \\
Qwen2.5-14B              & 3,362 &  726 &  411 & 1,880 & 1,166 & 15 \\
Meta-Llama-3-8B-Instruct & 3,118 &  640 &  405 & 1,394 & 1,305 & 34 \\
\bottomrule
\end{tabular}
\vspace{-2mm}
\label{tab:forward_model}
\end{table}


\begin{table}[t]
\footnotesize
\centering
\setlength{\tabcolsep}{0.08cm}
\caption{Top-10 models sorted by backward subgraph size.}
\vspace{-4mm}
\begin{tabular}{l c r r r r l}
\toprule
\textbf{Model} & 
\textbf{\begin{tabular}[c]{@{}c@{}}Model\\ Type\end{tabular}} & 
\textbf{Total} & 
\textbf{\begin{tabular}[c]{@{}r@{}}Fine-\\ tune\end{tabular}} & 
\textbf{\begin{tabular}[c]{@{}r@{}}Quanti-\\ zation\end{tabular}} & 
\textbf{Level} & 
\textbf{Base Model} \\
\midrule
command-r-1-layer    & Finetune & 40 & 39 & 0  & 39 & c4ai \\
KoModernBERT         & Finetune & 21 & 20 & 0  & 20 & ModernBERT \\
t5-small             & Finetune & 21 & 20 & 0  & 20 & t5-small \\
clinical\_260k       & Finetune & 20 & 19 & 0  & 19 & clinical\_180K \\
t5-small-finetuned   & Finetune & 17 & 16 & 0  & 16 & t5-small \\
clinical\_300k       & Finetune & 16 & 15 & 0  & 15 & clinical\_180K \\
clinical\_259k       & Finetune & 16 & 15 & 0  & 15 & clinical\_180K \\
LeoPARD-0.8.1        & Finetune & 16 &  2 & 13 & 15 & DeepSeek-R1 \\
LeoPARD-0.8.2-4bit   & Quantization & 16 &  1 & 14 & 15 & DeepSeek-R1 \\
LeoPARD-0.8.1-4bit   & Quantization & 16 &  1 & 14 & 15 & DeepSeek-R1 \\
\bottomrule
\end{tabular}
\vspace{-4mm}
\label{tab:backward_model}
\end{table}

\textbf{Task-specific model analysis.}
Given a task-specific model, we want to understand how it evolves, e.g., what other models it relies on. To achieve that, for each model, we perform a backward analysis by running BFS following the incoming edges. To that end, the derived subgraph shows the models it relies on.

{Table~\ref{tab:backward_model} shows the top-10 task-specific models sorted by the backward subgraph size, which is the number of models they rely on. We make two interesting observations.} 
(i) A fine-tuned model, \textit{command-r-1-layer}, illustrates the depth and complexity of transformations in the LLM supply chain. This model operates in bfloat16 (BF16) precision for efficient text generation and natural language understanding~\cite{pignatelli2024assessing}, originates from the base model \textit{c4ai}, and has undergone extensive lineage evolution before reaching its final form. Specifically, it depends on 40 upstream artifacts, including 39 other fine-tuned models, and spans 39 transformation levels in its backward lineage chain, as detailed in Table~\ref{tab:backward_model}. 
\update{(ii) We observe that adapters are mainly used for lightweight fine-tuning and merges for model integration, but task-specific models like \textit{command-r-1-layer}, as optimized standalone derivatives, do not evolve from adapters or merges in their backward lineage~\cite{pignatelli2024assessing}.}

\vspace{1mm}
\noindent{\textbf{Finding \#3:}
\textit{Base models like \textit{Llama-3.1-8B} dominate the LLM supply chain, spawning thousands of derivatives, while task-specific models such as \textit{command-r-1-layer} exhibit deep dependencies with other task-specific variants but avoid adapters or merges.}

\subsection{RQ \#4: Supply Chain of Models and Datasets}

\noindent This research question explores the interconnections between models and datasets 
from dual perspectives, including one dataset versus multiple models and one model versus multiple datasets.

{\textbf{One dataset versus multiple models} refers to the case
when a single dataset is used to train multiple models.}
Table~\ref{tab:dataset_model_1} shows the top-10 datasets based on the number of models trained on them.
In particular, 
(i) \textit{Mistral-v0.1} takes the leading position and is a widely adopted open-source dataset known for its strong performance in general-purpose language understanding and generation tasks. It has been used to train 1,093 models, including 300 fine-tuned variants, 300 adapters, 193 quantized models, and 300 merged models, highlighting its broad adoption across diverse model derivation strategies. 
\begin{table}[t]
\centering
\footnotesize  
\setlength{\tabcolsep}{0.1cm}
\caption{Top-10 datasets sorted by \# of models trained.}
\vspace{-4mm}
\begin{tabularx}{\columnwidth}{l r r r r r}  
\toprule
\textbf{Dataset} & 
\multicolumn{1}{r}{\textbf{Total}} & 
\textbf{Fine-tune} & 
\textbf{Adapter} & 
\textbf{Quantization} & 
\multicolumn{1}{r}{\textbf{Merge}} \\
\midrule
Mistral-v0.1        & 1,093 & 300 & 300 & 193 & 300 \\
TinyLlama-1.1B-v1.0    &  728 & 300 & 300 & 100 &  28 \\
open\_llama\_3b        &  304 &  15 & 285 &   4 &   0 \\
Yarn-Mistral-7b-128k   &  301 &   8 & 279 &  14 &   0 \\
WizardVicuna-open-llama&  280 &  12 & 261 &   7 &   0 \\
TinyLlama-1.1B-v0.6    &  266 &  10 & 243 &  13 &   0 \\
Yarn-Mistral-7b-64k    &  248 &   0 & 242 &   6 &   0 \\
Nous-Capybara-7B-V1    &  213 &  11 & 174 &  27 &   1 \\
MAmmoTH2-7B            &  213 &   0 &   0 &   3 & 210 \\
Starling-LM-7B-alpha   &  210 &  10 & 165 &  18 &  17 \\
\bottomrule
\end{tabularx}
\label{tab:dataset_model_1}
\vspace{-3mm}
\end{table}
(ii) The dataset \textit{TinyLlama-1.1B-v1.0} has been used to train 728 models, featuring 300 fine-tuned variants and 300 adapters. Similarly, \textit{open\_llama\_3b}, an open-access dataset of Llama, supports 285 adapter-based models, indicating a preference for lightweight, modular adaptation. 
(iii) Further, \textit{MAmmoTH2-7B} stands out with 210 merged models, showcasing its role in ensemble-style model fusion rather than traditional fine-tuning or adapter strategies. 

\textbf{One model versus multiple datasets} refers to the case when an LLM model is trained with multiple datasets.
Table~\ref{tab:dataset_model_2} shows the top-10 models ranked by the number of datasets used for training.
We observe that \textit{DeBERTa-ST-AllLayers-v3.1}, a fine-tuned variant of the \textit{DeBERTa} model, takes the top position, having been trained on 116 different datasets. Its adapter-based counterpart, \textit{DeBERTa-ST-AllLayers-v3.1bis}, also leverages the same number of datasets via adapter-based training, emphasizing modular reuse across tasks.

\begin{table}[t]
\footnotesize
\centering
\setlength{\tabcolsep}{2.2mm}
\caption{Top-10 models sorted by \# of training datasets.
}
\vspace{-4mm}
\begin{tabular}{llr}
\toprule
\textbf{Model} & 
\multicolumn{1}{r}{\textbf{\begin{tabular}[c]{@{}r@{}}Model Type \end{tabular}}} & 
\multicolumn{1}{r}{\textbf{\begin{tabular}[c]{@{}r@{}}\# of training dataset\end{tabular}}} \\
\midrule
DeBERTa-ST-AllLayers-v3.1 & Fine tune & 116 \\
DeBERTa-ST-AllLayers-v3.1bis & Adapters & 116 \\
static-similarity-mrl-mul-v1 & Fine tune & 108 \\
static-similarity-mrl-multilingual & Fine tune & 108 \\
ModernBERT-base-embed & Fine tune & 88 \\
Llama-3.2-3B-Instruct & Fine tune & 87 \\
Llama-3.2-3B-Instruct-GGUF & Quantization & 87 \\
DavidLanz-3.2-3B-Instruct & Fine tune & 87 \\
static-retrieval-mrl-en-v1 & Fine tune & 79 \\
XLMRoBERTaM3-CustomPoolin & Fine tune  & 72 \\
\bottomrule
\label{tab:dataset_model_2}
\vspace{-4mm}
\end{tabular}
\end{table}

\begin{figure}[t]
  \centering
    \vspace{1mm}
    \includegraphics[width=0.46\textwidth]{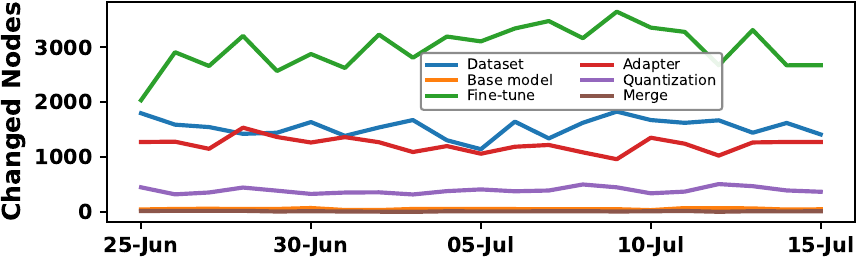} 
    \vspace{-2mm}
    \caption{The number of changed datasets and models on Hugging Face from June 25 to July 15, 2025.}
    \vspace{-4mm}
  \label{fig:daily_delta}
\end{figure}

In addition, models like \textit{static-similarity-mrl-mul-v1} and \textit{static-similarity-mrl-multilingual} are both fine-tuned on 108 datasets, indicating their roles in multilingual and multi-task similarity-based retrieval applications. 
Interestingly, most of these models are fine-tuned, specifically, 8 out of the top-10, 
suggesting that fine-tuning remains a dominant strategy for adapting base models to downstream tasks across heterogeneous data sources.

\vspace{1mm}
\noindent\textbf{Finding \#4:}
\textit{Models and datasets exhibit strong bidirectional interdependence, with datasets like \textit{Mistral-v0.1} spawning hundreds of models, while models such as \textit{DeBERTa-ST-AllLayers-v3.1} leverage different datasets to enhance adaptability, highlighting the critical roles of dataset-model interactions in advancing AI.}

\subsection{RQ \#5: Dynamic Update Evaluation}
\label{sec:graph_updates}

\noindent This research question aims to understand the dynamic update of the LLM supply chain.
We perform a daily-based data collection by capturing the addition and deletion of nodes and edges each day.
Figure~\ref{fig:daily_delta} denotes the number of changed (additions and deletions) datasets and models 
from June 25 to July 15, 2025.
We make three interesting observations.
(i) The daily dynamic update is significant. That is, an average of 4,622 models are changing every day, including 
$\sim$3,843 model additions and $\sim$779 deletions.
In addition, about 1,538 datasets are changing each day, containing  
$\sim$1,305 dataset additions and $\sim$233 deletions.
\update{An addition occurs when a new model or dataset is uploaded to the Hugging Face platform. This includes base models, task-specific variants, and new training datasets. A deletion refers to the removal of such nodes, often due to licensing issues, privacy concerns, or contributor decisions, such as replacing outdated models or withdrawing low-quality or sensitive datasets.}


(ii) Fine-tuned models dominate daily activity, averaging over \update{2,988 changes per day, followed by consistent contributions from adapters ($\sim$1,224/day) and datasets ($\sim$1,539/day).} Noticeable spikes, such as on July 7 and July 9, align with major events like the \textit{Mistral-Fusion-v3} fine-tuning wave and dataset updates such as \textit{HFTime2025-News}. (iii) Furthermore, adapter uploads peaked at 1,249 on June 28, while quantized variants reached 381 on July 9, driven by releases such as \textit{Qwen-GGUF-7B}. These patterns demonstrate {\name}'s ability to capture evolving supply chain dynamics at a fine-grained level.





\vspace{1mm}
\noindent\textbf{Finding \#5:}
\textit{The LLM supply chain exhibits continuous and high-volume daily changes, driven by frequent additions and deletions of models and datasets. This reflects a rapidly evolving and highly dynamic ecosystem shaped by active contributor behavior.}

\vspace{-1mm}
\subsection{RQ \#6: Generalization to Other AI Platforms}
\label{sec:kaggle_results}

\update{To validate {\name}’s generalizability beyond Hugging Face, we applied our pipeline to another AI platform, Kaggle~\cite{bojer2021kaggle}. 
As of July 25, 2025, Kaggle hosts 470 base models, 3,146 task-specific models, and $\sim$502K datasets. 
Using Kaggle's kernel and dataset APIs, we collected 2,640 models and 105,867 datasets. This significant gap is due to a lack of models' and datasets' metadata. Of the datasets retrieved ($\sim$105K), only 137 datasets were included in our graph, as most lacked standardized documentation or traceable links to models. Many are standalone, poorly described, or lack contextual information, a challenge also observed on Hugging Face.}


\update{We follow the same way to construct a heterogeneous graph consisting of 2,777 nodes, which include 2,640 model nodes, comprising 59 base models, 2,410 fine-tuned, 171 quantization models, and 137 dataset nodes. The graph contains a total of 3,990 edges, on which we observed seven types of edges, (i) base model $\rightarrow$ fine-tuned model (467 edges), (ii) base model $\rightarrow$ quantization model (62 edges), (iii) fine-tuned model $\rightarrow$ fine-tuned model (1,696 edges), (iv) fine-tuned model $\rightarrow$ quantized model (107 edges), (v) quantized model $\rightarrow$ quantized model (1 edge), (vi) dataset $\rightarrow$ fine-tuned model (1,614 edges), and (vii) dataset $\rightarrow$ quantization model (43 edges).}

\update{We observed that the average degree is 1.44, indicating that the graph is sparse. 
We made two interesting observations.
(i) The degree distribution is heavy-tailed and skewed: out of 2,777 total nodes, 2,305 nodes~(83\%) have a total degree of 1.}
\update{Most nodes have low degrees, while a few highly connected hubs dominate the graph. For example, in-degrees range from 0 to 4,
and \textit{tensorflow/mobilenet-v1} has the highest out-degree of 64, followed by \textit{google/nnlm} with 56.
(ii) Furthermore, the graph contains 448 WCCs, reflecting high fragmentation. However, the largest WCC includes 65 nodes, suggesting the presence of a moderately sized core subgraph.}



\vspace{1mm}
\noindent\textbf{Finding \#6:}
\textit{The resulting graph exhibits structural properties consistent with our Hugging Face analysis, including a heavy-tailed degree distribution, sparse connectivity, and strong modular fragmentation, demonstrating the robustness and generalizability of our pipeline across platforms despite metadata limitations.}
\vspace{-1mm}
\section{Use Case}

\noindent{{\name} presents a technique to analyze the supply chain of the LLM ecosystem},
which can be used for various applications, e.g., auditing provenance, identifying biases, and revealing trends like quantized model scarcity.
We discuss the two use cases.


\textbf{Use case \#1: Tracing lineage and dependency in LLM supply chain.}
Models are frequently built upon others through fine-tuning, adapter training, or quantization, forming complex chains of dependencies. However, when these relationships are not explicitly visible, it becomes difficult to verify where a model comes from, whether it inherits bias from upstream datasets, or if it complies with licensing constraints. {\name} can be used to address this challenge by constructing the supply chain of models and datasets, uncovering both direct and derived dependencies, even when they are not formally documented. 
For example, it can trace how the model \textit{Meta-llama} indirectly relies on a dataset like \textit{Wikimedia} via \textit{Chatgpt-prompt} (Figure~\ref{fig:sample_subgraph}). This transparency supports developers, auditors, and policymakers in validating provenance, detecting risks, and enabling trustworthy AI.

\textbf{Use case \#2: Identifying critical nodes and structural vulnerabilities.}
In the LLM ecosystem, certain models (e.g., \textit{gemma-2b}) and datasets (e.g., \textit{The Pile}) are reused so frequently that they become critical structural hubs, where their failure or removal could disrupt numerous downstream dependencies. These hidden single points of failure are difficult to detect without a comprehensive view of resource interconnections. {\name} can be used to address this by modeling the supply chain as a graph and analyzing node connectivity to surface highly reused models and datasets with significant inbound or outbound links. This visibility enables maintainers to safeguard vital assets and helps developers mitigate the risk of overreliance on fragile or under-maintained components.


\section{Related Work and Discussion}
\label{sec: related work}

\textbf{LLM supply chain perspectives in AI.} LLMs are advancing across model infrastructure, lifecycle, and applications~\cite{wang2024large}. Model reuse is widespread, promoting large-scale sharing and adaptation of base models~\cite{jiang2023empirical}. Open-source ecosystems such as Hugging Face host diverse LLMs and datasets, democratizing AI~\cite{riva2024huggingface}. Base models~\cite{xu2024survey}, trained on broad datasets, enable task-specific variants via fine-tuning~\cite{han2021pre}, reinforcing democratization and innovation~\cite{cupac2024democratization}.

\textbf{Relationship analysis between LLM models and datasets.} A recent study investigates the practical adaptation of base models to specific tasks. Multitask fine-tuning has demonstrated the potential to enhance performance on target tasks with scarce labels~\cite{xu2024towards}. In plant phenotyping, adapting vision-based models by techniques like adapter tuning and decoder tuning has shown results comparable to those of leading task-specific models~\cite{chen2023adapting}. 
The Quadapter technique for language models tackles quantization difficulties by incorporating learnable parameters that scale activations channel-wise, mitigating overfitting during quantization-aware training~\cite{park2022quadapter}.  

In future, we would like to explore more through the following perspectives,
(i) collecting more LLM supply chain information accurately and scalably;
(ii) understanding the LLM supply chain better via both fundamental graph analytics~\cite{feng2023peek,hu2025adaptive,li2025optimized} and graph AI techniques~\cite{kipf2016semi,fu2022tlpgnn,wang2025bingo,ji2025hpcgraphai};
and (iii) exploring the security and privacy threats on the LLM supply chain~\cite{ji2021buggraph,he2024code,ji2021definit,cui2023api2vec,brucker2024cloudcover}.

\vspace{-1mm}

\section{Conclusion}
\label{conclusion}

\noindent{This project studies the relationships} between models and datasets in the LLM ecosystem, which are the central parts of the \textit{LLM supply chain}.
First, we systematically collect the supply chain information of LLMs. 
With that, we construct a directed heterogeneous graph, having \update{\textit{402,654 nodes} and \textit{462,524 edges}}.
Lastly, we perform different types of analysis and make multiple interesting findings.
\vspace{-1mm}
\section*{Acknowledgment}
\noindent This work was supported in part by National Science Foundation grants 2331301, 2508118, 2516003, 2419843.
The views, opinions, and/or findings expressed in this material are those of the authors and should not be interpreted as representing the official views of the National Science Foundation, or the U.S. Government.

\section*{Disclaimer on Generative AI Usage}
\noindent This work does not use generative AI tools (e.g., ChatGPT, Copilot, Bard, Claude, etc.) for ideation, analysis, code development, or writing. All content, experiments, and results presented in this paper were created and validated solely by the authors. Mentions of generative AI are included only for scholarly context or comparison.


\begin{thebibliography}{62}


\ifx \showCODEN    \undefined \def \showCODEN     #1{\unskip}     \fi
\ifx \showISBNx    \undefined \def \showISBNx     #1{\unskip}     \fi
\ifx \showISBNxiii \undefined \def \showISBNxiii  #1{\unskip}     \fi
\ifx \showISSN     \undefined \def \showISSN      #1{\unskip}     \fi
\ifx \showLCCN     \undefined \def \showLCCN      #1{\unskip}     \fi
\ifx \shownote     \undefined \def \shownote      #1{#1}          \fi
\ifx \showarticletitle \undefined \def \showarticletitle #1{#1}   \fi
\ifx \showURL      \undefined \def \showURL       {\relax}        \fi
\providecommand\bibfield[2]{#2}
\providecommand\bibinfo[2]{#2}
\providecommand\natexlab[1]{#1}
\providecommand\showeprint[2][]{arXiv:#2}

\bibitem[Agossah et~al\mbox{.}(2023)]%
        {agossah2023llm}
\bibfield{author}{\bibinfo{person}{Alexandre Agossah}, \bibinfo{person}{Fr{\'e}d{\'e}rique Krupa}, \bibinfo{person}{Matthieu Perreira Da~Silva}, {and} \bibinfo{person}{Patrick Le~Callet}.} \bibinfo{year}{2023}\natexlab{}.
\newblock \showarticletitle{Llm-based interaction for content generation: A case study on the perception of employees in an it department}. In \bibinfo{booktitle}{\emph{Proceedings of the 2023 ACM International Conference on Interactive Media Experiences}}. \bibinfo{pages}{237--241}.
\newblock


\bibitem[Akiba et~al\mbox{.}(2025)]%
        {akiba2025evolutionary}
\bibfield{author}{\bibinfo{person}{Takuya Akiba}, \bibinfo{person}{Makoto Shing}, \bibinfo{person}{Yujin Tang}, \bibinfo{person}{Qi Sun}, {and} \bibinfo{person}{David Ha}.} \bibinfo{year}{2025}\natexlab{}.
\newblock \showarticletitle{Evolutionary optimization of model merging recipes}.
\newblock \bibinfo{journal}{\emph{Nature Machine Intelligence}} \bibinfo{volume}{7}, \bibinfo{number}{2} (\bibinfo{year}{2025}), \bibinfo{pages}{195--204}.
\newblock


\bibitem[Allemang and Sequeda(2024)]%
        {allemang2024increasing}
\bibfield{author}{\bibinfo{person}{Dean Allemang} {and} \bibinfo{person}{Juan Sequeda}.} \bibinfo{year}{2024}\natexlab{}.
\newblock \showarticletitle{Increasing the LLM Accuracy for Question Answering: Ontologies to the Rescue!}
\newblock \bibinfo{journal}{\emph{arXiv preprint arXiv:2405.11706}} (\bibinfo{year}{2024}).
\newblock


\bibitem[Anonymous(2022)]%
        {manufacturing2022review}
\bibfield{author}{\bibinfo{person}{Anonymous}.} \bibinfo{year}{2022}\natexlab{}.
\newblock \showarticletitle{Review of Supply Chain Management in Manufacturing Organizations}.
\newblock \bibinfo{journal}{\emph{ResearchGate}} (\bibinfo{year}{2022}).
\newblock
\urldef\tempurl%
\url{https://www.researchgate.net/publication/377659033_Review_of_supply_chain_management_in_manufacturing_organizations}
\showURL{%
\tempurl}


\bibitem[Bahani et~al\mbox{.}(2023)]%
        {bahani2023effectiveness}
\bibfield{author}{\bibinfo{person}{Mourad Bahani}, \bibinfo{person}{Aziza El~Ouaazizi}, {and} \bibinfo{person}{Khalil Maalmi}.} \bibinfo{year}{2023}\natexlab{}.
\newblock \showarticletitle{The effectiveness of T5, GPT-2, and BERT on text-to-image generation task}.
\newblock \bibinfo{journal}{\emph{Pattern Recognition Letters}}  \bibinfo{volume}{173} (\bibinfo{year}{2023}), \bibinfo{pages}{57--63}.
\newblock


\bibitem[Bojer and Meldgaard(2021)]%
        {bojer2021kaggle}
\bibfield{author}{\bibinfo{person}{Casper~Solheim Bojer} {and} \bibinfo{person}{Jens~Peder Meldgaard}.} \bibinfo{year}{2021}\natexlab{}.
\newblock \showarticletitle{Kaggle forecasting competitions: An overlooked learning opportunity}.
\newblock \bibinfo{journal}{\emph{International Journal of Forecasting}} \bibinfo{volume}{37}, \bibinfo{number}{2} (\bibinfo{year}{2021}), \bibinfo{pages}{587--603}.
\newblock


\bibitem[Bonner et~al\mbox{.}(2023)]%
        {bonner2023large}
\bibfield{author}{\bibinfo{person}{Euan Bonner}, \bibinfo{person}{Ryan Lege}, {and} \bibinfo{person}{Erin Frazier}.} \bibinfo{year}{2023}\natexlab{}.
\newblock \showarticletitle{Large Language Model-Based Artificial Intelligence in the Language Classroom: Practical Ideas for Teaching.}
\newblock \bibinfo{journal}{\emph{Teaching English with Technology}} \bibinfo{volume}{23}, \bibinfo{number}{1} (\bibinfo{year}{2023}), \bibinfo{pages}{23--41}.
\newblock


\bibitem[Brucker-Hahn et~al\mbox{.}(2024)]%
        {brucker2024cloudcover}
\bibfield{author}{\bibinfo{person}{Dalton~A Brucker-Hahn}, \bibinfo{person}{Wang Feng}, \bibinfo{person}{Shanchao Li}, \bibinfo{person}{Matthew Petillo}, \bibinfo{person}{Alexandru~G Bardas}, \bibinfo{person}{Drew Davidson}, {and} \bibinfo{person}{Yuede Ji}.} \bibinfo{year}{2024}\natexlab{}.
\newblock \showarticletitle{CloudCover: Enforcement of Multi-Hop Network Connections in Microservice Deployments}. In \bibinfo{booktitle}{\emph{2024 Annual Computer Security Applications Conference (ACSAC)}}. IEEE, \bibinfo{pages}{1186--1202}.
\newblock


\bibitem[Chen et~al\mbox{.}(2023)]%
        {chen2023adapting}
\bibfield{author}{\bibinfo{person}{Feng Chen}, \bibinfo{person}{Mario~Valerio Giuffrida}, {and} \bibinfo{person}{Sotirios~A Tsaftaris}.} \bibinfo{year}{2023}\natexlab{}.
\newblock \showarticletitle{Adapting vision foundation models for plant phenotyping}. In \bibinfo{booktitle}{\emph{Proceedings of the IEEE/CVF International Conference on Computer Vision}}. \bibinfo{pages}{604--613}.
\newblock


\bibitem[Chou et~al\mbox{.}(2006)]%
        {chou2006software}
\bibfield{author}{\bibinfo{person}{M.~C. Chou}, \bibinfo{person}{H. Ye}, \bibinfo{person}{X.~M. Yuan}, \bibinfo{person}{Y.~N. Cheng}, \bibinfo{person}{L. Chua}, \bibinfo{person}{Y. Guan}, \bibinfo{person}{S.~E. Lee}, {and} \bibinfo{person}{Y.~C. Tay}.} \bibinfo{year}{2006}\natexlab{}.
\newblock \showarticletitle{Analysis of a Software-Focused Products and Service Supply Chain}.
\newblock \bibinfo{journal}{\emph{IEEE Transactions on Industrial Informatics}} \bibinfo{volume}{2}, \bibinfo{number}{4} (\bibinfo{year}{2006}), \bibinfo{pages}{295--303}.
\newblock
\href{https://doi.org/10.1109/TII.2006.884368}{doi:\nolinkurl{10.1109/TII.2006.884368}}


\bibitem[Cui et~al\mbox{.}(2023)]%
        {cui2023api2vec}
\bibfield{author}{\bibinfo{person}{Lei Cui}, \bibinfo{person}{Jiancong Cui}, \bibinfo{person}{Yuede Ji}, \bibinfo{person}{Zhiyu Hao}, \bibinfo{person}{Lun Li}, {and} \bibinfo{person}{Zhenquan Ding}.} \bibinfo{year}{2023}\natexlab{}.
\newblock \showarticletitle{API2Vec: Learning Representations of API Sequences for Malware Detection}. In \bibinfo{booktitle}{\emph{International Symposium on Software Testing and Analysis (ISSTA)}}.
\newblock


\bibitem[Cupa{\'c} et~al\mbox{.}(2024)]%
        {cupac2024democratization}
\bibfield{author}{\bibinfo{person}{Jelena Cupa{\'c}}, \bibinfo{person}{Hendrik Schopmans}, {and} \bibinfo{person}{{\.I}rem Tuncer-Ebet{\"u}rk}.} \bibinfo{year}{2024}\natexlab{}.
\newblock \bibinfo{title}{Democratization in the age of artificial intelligence: introduction to the special issue}.
\newblock \bibinfo{numpages}{899--921}~pages.
\newblock


\bibitem[De~Meo et~al\mbox{.}(2011)]%
        {de2011generalized}
\bibfield{author}{\bibinfo{person}{Pasquale De~Meo}, \bibinfo{person}{Emilio Ferrara}, \bibinfo{person}{Giacomo Fiumara}, {and} \bibinfo{person}{Alessandro Provetti}.} \bibinfo{year}{2011}\natexlab{}.
\newblock \showarticletitle{Generalized louvain method for community detection in large networks}. In \bibinfo{booktitle}{\emph{2011 11th international conference on intelligent systems design and applications}}. IEEE, \bibinfo{pages}{88--93}.
\newblock


\bibitem[Face({[n.\,d.]})]%
        {huggingface}
\bibfield{author}{\bibinfo{person}{Hugging Face}.} \bibinfo{year}{[n.\,d.]}\natexlab{}.
\newblock \bibinfo{title}{{Hugging Face -- The AI community building the future}}.
\newblock
\urldef\tempurl%
\url{https://huggingface.co/}
\showURL{%
\tempurl}


\bibitem[Feng et~al\mbox{.}(2023)]%
        {feng2023peek}
\bibfield{author}{\bibinfo{person}{Wang Feng}, \bibinfo{person}{Shiyang Chen}, \bibinfo{person}{Hang Liu}, {and} \bibinfo{person}{Yuede Ji}.} \bibinfo{year}{2023}\natexlab{}.
\newblock \showarticletitle{Peek: A Prune-Centric Approach for K Shortest Path Computation}. In \bibinfo{booktitle}{\emph{Proceedings of the International Conference for High Performance Computing, Networking, Storage and Analysis}}. \bibinfo{pages}{1--14}.
\newblock


\bibitem[Fu et~al\mbox{.}(2022)]%
        {fu2022tlpgnn}
\bibfield{author}{\bibinfo{person}{Qiang Fu}, \bibinfo{person}{Yuede Ji}, {and} \bibinfo{person}{H~Howie Huang}.} \bibinfo{year}{2022}\natexlab{}.
\newblock \showarticletitle{TLPGNN: A lightweight two-level parallelism paradigm for graph neural network computation on GPU}. In \bibinfo{booktitle}{\emph{HPDC}}.
\newblock


\bibitem[Grattafiori et~al\mbox{.}(2024)]%
        {grattafiori2024llama}
\bibfield{author}{\bibinfo{person}{Aaron Grattafiori}, \bibinfo{person}{Abhimanyu Dubey}, \bibinfo{person}{Abhinav Jauhri}, \bibinfo{person}{Abhinav Pandey}, \bibinfo{person}{Abhishek Kadian}, \bibinfo{person}{Ahmad Al-Dahle}, \bibinfo{person}{Aiesha Letman}, \bibinfo{person}{Akhil Mathur}, \bibinfo{person}{Alan Schelten}, \bibinfo{person}{Alex Vaughan}, {et~al\mbox{.}}} \bibinfo{year}{2024}\natexlab{}.
\newblock \showarticletitle{The llama 3 herd of models}.
\newblock \bibinfo{journal}{\emph{arXiv preprint arXiv:2407.21783}} (\bibinfo{year}{2024}).
\newblock


\bibitem[Han et~al\mbox{.}(2021)]%
        {han2021pre}
\bibfield{author}{\bibinfo{person}{Xu Han}, \bibinfo{person}{Zhengyan Zhang}, \bibinfo{person}{Ning Ding}, \bibinfo{person}{Yuxian Gu}, \bibinfo{person}{Xiao Liu}, \bibinfo{person}{Yuqi Huo}, \bibinfo{person}{Jiezhong Qiu}, \bibinfo{person}{Yuan Yao}, \bibinfo{person}{Ao Zhang}, \bibinfo{person}{Liang Zhang}, {et~al\mbox{.}}} \bibinfo{year}{2021}\natexlab{}.
\newblock \showarticletitle{Pre-trained models: Past, present and future}.
\newblock \bibinfo{journal}{\emph{AI Open}}  \bibinfo{volume}{2} (\bibinfo{year}{2021}), \bibinfo{pages}{225--250}.
\newblock


\bibitem[He et~al\mbox{.}(2024a)]%
        {he2024code}
\bibfield{author}{\bibinfo{person}{Haojie He}, \bibinfo{person}{Xingwei Lin}, \bibinfo{person}{Ziang Weng}, \bibinfo{person}{Ruijie Zhao}, \bibinfo{person}{Shuitao Gan}, \bibinfo{person}{Libo Chen}, \bibinfo{person}{Yuede Ji}, \bibinfo{person}{Jiashui Wang}, {and} \bibinfo{person}{Zhi Xue}.} \bibinfo{year}{2024}\natexlab{a}.
\newblock \showarticletitle{Code is not Natural Language: Unlock the Power of Semantics-Oriented Graph Representation for Binary Code Similarity Detection}. In \bibinfo{booktitle}{\emph{The 33rd USENIX Security Symposium (USENIX Security)}}.
\newblock


\bibitem[He et~al\mbox{.}(2024b)]%
        {he2024llama}
\bibfield{author}{\bibinfo{person}{Zhengfu He}, \bibinfo{person}{Wentao Shu}, \bibinfo{person}{Xuyang Ge}, \bibinfo{person}{Lingjie Chen}, \bibinfo{person}{Junxuan Wang}, \bibinfo{person}{Yunhua Zhou}, \bibinfo{person}{Frances Liu}, \bibinfo{person}{Qipeng Guo}, \bibinfo{person}{Xuanjing Huang}, \bibinfo{person}{Zuxuan Wu}, {et~al\mbox{.}}} \bibinfo{year}{2024}\natexlab{b}.
\newblock \showarticletitle{Llama scope: Extracting millions of features from llama-3.1-8b with sparse autoencoders}.
\newblock \bibinfo{journal}{\emph{arXiv preprint arXiv:2410.20526}} (\bibinfo{year}{2024}).
\newblock


\bibitem[Hu et~al\mbox{.}(2025)]%
        {hu2025adaptive}
\bibfield{author}{\bibinfo{person}{Runbang Hu}, \bibinfo{person}{Chaoqun Li}, \bibinfo{person}{Xiaojiang Du}, {and} \bibinfo{person}{Yuede Ji}.} \bibinfo{year}{2025}\natexlab{}.
\newblock \showarticletitle{Adaptive Optimizations for Parallel Single-Source Shortest Paths}.
\newblock In \bibinfo{booktitle}{\emph{Proceedings of the 1st FastCode Programming Challenge}}. \bibinfo{pages}{53--56}.
\newblock


\bibitem[Hu et~al\mbox{.}(2023)]%
        {hu2023llm}
\bibfield{author}{\bibinfo{person}{Zhiqiang Hu}, \bibinfo{person}{Lei Wang}, \bibinfo{person}{Yihuai Lan}, \bibinfo{person}{Wanyu Xu}, \bibinfo{person}{Ee-Peng Lim}, \bibinfo{person}{Lidong Bing}, \bibinfo{person}{Xing Xu}, \bibinfo{person}{Soujanya Poria}, {and} \bibinfo{person}{Roy Ka-Wei Lee}.} \bibinfo{year}{2023}\natexlab{}.
\newblock \showarticletitle{Llm-adapters: An adapter family for parameter-efficient fine-tuning of large language models}.
\newblock \bibinfo{journal}{\emph{arXiv preprint arXiv:2304.01933}} (\bibinfo{year}{2023}).
\newblock


\bibitem[Ji(2025)]%
        {ji2025hpcgraphai}
\bibfield{author}{\bibinfo{person}{Yuede Ji}.} \bibinfo{year}{2025}\natexlab{}.
\newblock \showarticletitle{High-Performance Computing for Graph AI: A Top-Down Perspective}. In \bibinfo{booktitle}{\emph{Proceedings of the 15th NSF/TCPP Workshop on Parallel and Distributed Computing Education (EduPar '25)}}.
\newblock


\bibitem[Ji et~al\mbox{.}(2021a)]%
        {ji2021buggraph}
\bibfield{author}{\bibinfo{person}{Yuede Ji}, \bibinfo{person}{Lei Cui}, {and} \bibinfo{person}{H.~Howie Huang}.} \bibinfo{year}{2021}\natexlab{a}.
\newblock \showarticletitle{{BugGraph: Differentiating Source-Binary Code Similarity with Graph Triplet-Loss Network}}. In \bibinfo{booktitle}{\emph{16th ACM ASIA Conference on Computer and Communications Security (AsiaCCS)}}.
\newblock


\bibitem[Ji et~al\mbox{.}(2021b)]%
        {ji2021definit}
\bibfield{author}{\bibinfo{person}{Yuede Ji}, \bibinfo{person}{Mohamed Elsabagh}, \bibinfo{person}{Ryan Johnson}, {and} \bibinfo{person}{Angelos Stavrou}.} \bibinfo{year}{2021}\natexlab{b}.
\newblock \showarticletitle{{DEFInit: An Analysis of Exposed Android Init Routines}}. In \bibinfo{booktitle}{\emph{30th USENIX Security Symposium (USENIX Security)}}.
\newblock


\bibitem[Ji and Huang(2020)]%
        {ji2020aquila}
\bibfield{author}{\bibinfo{person}{Yuede Ji} {and} \bibinfo{person}{H.~Howie Huang}.} \bibinfo{year}{2020}\natexlab{}.
\newblock \showarticletitle{{Aquila: Adaptive Parallel Computation of Graph Connectivity Queries}}. In \bibinfo{booktitle}{\emph{Proceedings of the 29th International Symposium on High-Performance Parallel and Distributed Computing (HPDC)}}.
\newblock


\bibitem[Ji et~al\mbox{.}(2018)]%
        {ji2018ispan}
\bibfield{author}{\bibinfo{person}{Yuede Ji}, \bibinfo{person}{Hang Liu}, {and} \bibinfo{person}{H.~Howie Huang}.} \bibinfo{year}{2018}\natexlab{}.
\newblock \showarticletitle{{iSpan: Parallel Identification of Strongly Connected Components with Spanning Trees}}. In \bibinfo{booktitle}{\emph{International Conference for High Performance Computing, Networking, Storage and Analysis (SC)}}. IEEE, \bibinfo{pages}{731--742}.
\newblock


\bibitem[Ji et~al\mbox{.}(2020)]%
        {ji2020swarmgraph}
\bibfield{author}{\bibinfo{person}{Yuede Ji}, \bibinfo{person}{Hang Liu}, {and} \bibinfo{person}{H.~Howie Huang}.} \bibinfo{year}{2020}\natexlab{}.
\newblock \showarticletitle{{SwarmGraph: Analyzing Large-Scale In-Memory Graphs on GPUs}}. In \bibinfo{booktitle}{\emph{International Conference on High Performance Computing and Communications (HPCC)}}. IEEE.
\newblock


\bibitem[Jiang et~al\mbox{.}(2023)]%
        {jiang2023empirical}
\bibfield{author}{\bibinfo{person}{Wenxin Jiang}, \bibinfo{person}{Nicholas Synovic}, \bibinfo{person}{Matt Hyatt}, \bibinfo{person}{Taylor~R Schorlemmer}, \bibinfo{person}{Rohan Sethi}, \bibinfo{person}{Yung-Hsiang Lu}, \bibinfo{person}{George~K Thiruvathukal}, {and} \bibinfo{person}{James~C Davis}.} \bibinfo{year}{2023}\natexlab{}.
\newblock \showarticletitle{An empirical study of pre-trained model reuse in the hugging face deep learning model registry}. In \bibinfo{booktitle}{\emph{2023 IEEE/ACM 45th International Conference on Software Engineering (ICSE)}}. IEEE, \bibinfo{pages}{2463--2475}.
\newblock


\bibitem[Kenton and Toutanova(2019)]%
        {kenton2019bert}
\bibfield{author}{\bibinfo{person}{Jacob Devlin Ming-Wei~Chang Kenton} {and} \bibinfo{person}{Lee~Kristina Toutanova}.} \bibinfo{year}{2019}\natexlab{}.
\newblock \showarticletitle{Bert: Pre-training of deep bidirectional transformers for language understanding}. In \bibinfo{booktitle}{\emph{Proceedings of naacL-HLT}}, Vol.~\bibinfo{volume}{1}. Minneapolis, Minnesota, \bibinfo{pages}{2}.
\newblock


\bibitem[Kipf(2016)]%
        {kipf2016semi}
\bibfield{author}{\bibinfo{person}{TN Kipf}.} \bibinfo{year}{2016}\natexlab{}.
\newblock \showarticletitle{Semi-Supervised Classification with Graph Convolutional Networks}.
\newblock \bibinfo{journal}{\emph{arXiv preprint arXiv:1609.02907}} (\bibinfo{year}{2016}).
\newblock


\bibitem[Knuth(1974)]%
        {Knuth1974}
\bibfield{author}{\bibinfo{person}{Donald~E. Knuth}.} \bibinfo{year}{1974}\natexlab{}.
\newblock \bibinfo{booktitle}{\emph{The Art of Computer Programming, Volume 1: Fundamental Algorithms} (\bibinfo{edition}{2nd} ed.)}.
\newblock \bibinfo{publisher}{Addison-Wesley}.
\newblock


\bibitem[Laban et~al\mbox{.}(2023)]%
        {laban2023summedits}
\bibfield{author}{\bibinfo{person}{Philippe Laban}, \bibinfo{person}{Wojciech Kry{\'s}ci{\'n}ski}, \bibinfo{person}{Divyansh Agarwal}, \bibinfo{person}{Alexander~Richard Fabbri}, \bibinfo{person}{Caiming Xiong}, \bibinfo{person}{Shafiq Joty}, {and} \bibinfo{person}{Chien-Sheng Wu}.} \bibinfo{year}{2023}\natexlab{}.
\newblock \showarticletitle{SUMMEDITS: measuring LLM ability at factual reasoning through the lens of summarization}. In \bibinfo{booktitle}{\emph{Proceedings of the 2023 Conference on Empirical Methods in Natural Language Processing}}. \bibinfo{pages}{9662--9676}.
\newblock


\bibitem[Li et~al\mbox{.}(2023)]%
        {li2023large}
\bibfield{author}{\bibinfo{person}{Beibin Li}, \bibinfo{person}{Konstantina Mellou}, \bibinfo{person}{Bo Zhang}, \bibinfo{person}{Jeevan Pathuri}, {and} \bibinfo{person}{Ishai Menache}.} \bibinfo{year}{2023}\natexlab{}.
\newblock \showarticletitle{Large language models for supply chain optimization}.
\newblock \bibinfo{journal}{\emph{arXiv preprint arXiv:2307.03875}} (\bibinfo{year}{2023}).
\newblock


\bibitem[Li et~al\mbox{.}(2025)]%
        {li2025optimized}
\bibfield{author}{\bibinfo{person}{Chaoqun Li}, \bibinfo{person}{Runbang Hu}, \bibinfo{person}{Xiaojiang Du}, {and} \bibinfo{person}{Yuede Ji}.} \bibinfo{year}{2025}\natexlab{}.
\newblock \showarticletitle{Optimized Parallel Breadth-First Search with Adaptive Strategies}.
\newblock In \bibinfo{booktitle}{\emph{Proceedings of the 1st FastCode Programming Challenge}}. \bibinfo{pages}{28--32}.
\newblock


\bibitem[Li et~al\mbox{.}(2020)]%
        {li2020survey}
\bibfield{author}{\bibinfo{person}{Jing Li}, \bibinfo{person}{Aixin Sun}, \bibinfo{person}{Jianglei Han}, {and} \bibinfo{person}{Chenliang Li}.} \bibinfo{year}{2020}\natexlab{}.
\newblock \showarticletitle{A survey on deep learning for named entity recognition}.
\newblock \bibinfo{journal}{\emph{IEEE transactions on knowledge and data engineering}} \bibinfo{volume}{34}, \bibinfo{number}{1} (\bibinfo{year}{2020}), \bibinfo{pages}{50--70}.
\newblock


\bibitem[Lu et~al\mbox{.}(2024)]%
        {lu2024llamax}
\bibfield{author}{\bibinfo{person}{Yinquan Lu}, \bibinfo{person}{Wenhao Zhu}, \bibinfo{person}{Lei Li}, \bibinfo{person}{Yu Qiao}, {and} \bibinfo{person}{Fei Yuan}.} \bibinfo{year}{2024}\natexlab{}.
\newblock \showarticletitle{Llamax: Scaling linguistic horizons of llm by enhancing translation capabilities beyond 100 languages}.
\newblock \bibinfo{journal}{\emph{arXiv preprint arXiv:2407.05975}} (\bibinfo{year}{2024}).
\newblock


\bibitem[Merrick et~al\mbox{.}(2024)]%
        {merrick2024upscaling}
\bibfield{author}{\bibinfo{person}{Felix Merrick}, \bibinfo{person}{Maria Radcliffe}, {and} \bibinfo{person}{Rupert Hensley}.} \bibinfo{year}{2024}\natexlab{}.
\newblock \showarticletitle{Upscaling a smaller llm to more parameters via manual regressive distillation}.
\newblock  (\bibinfo{year}{2024}).
\newblock


\bibitem[Mirchandani et~al\mbox{.}(2023)]%
        {mirchandani2023large}
\bibfield{author}{\bibinfo{person}{Suvir Mirchandani}, \bibinfo{person}{Fei Xia}, \bibinfo{person}{Pete Florence}, \bibinfo{person}{Brian Ichter}, \bibinfo{person}{Danny Driess}, \bibinfo{person}{Montserrat~Gonzalez Arenas}, \bibinfo{person}{Kanishka Rao}, \bibinfo{person}{Dorsa Sadigh}, {and} \bibinfo{person}{Andy Zeng}.} \bibinfo{year}{2023}\natexlab{}.
\newblock \showarticletitle{Large language models as general pattern machines}.
\newblock \bibinfo{journal}{\emph{arXiv preprint arXiv:2307.04721}} (\bibinfo{year}{2023}).
\newblock


\bibitem[Mizrahi et~al\mbox{.}(2024)]%
        {mizrahi2024state}
\bibfield{author}{\bibinfo{person}{Moran Mizrahi}, \bibinfo{person}{Guy Kaplan}, \bibinfo{person}{Dan Malkin}, \bibinfo{person}{Rotem Dror}, \bibinfo{person}{Dafna Shahaf}, {and} \bibinfo{person}{Gabriel Stanovsky}.} \bibinfo{year}{2024}\natexlab{}.
\newblock \showarticletitle{State of what art? a call for multi-prompt llm evaluation}.
\newblock \bibinfo{journal}{\emph{Transactions of the Association for Computational Linguistics}}  \bibinfo{volume}{12} (\bibinfo{year}{2024}), \bibinfo{pages}{933--949}.
\newblock


\bibitem[{ONNX Community}(2025)]%
        {onnx_model_zoo}
\bibfield{author}{\bibinfo{person}{{ONNX Community}}.} \bibinfo{year}{2025}\natexlab{}.
\newblock \bibinfo{title}{{ONNX Model Zoo}: Pre-trained Models for {ONNX}}.
\newblock \bibinfo{howpublished}{\url{https://github.com/onnx/models}}.
\newblock


\bibitem[Park et~al\mbox{.}(2022)]%
        {park2022quadapter}
\bibfield{author}{\bibinfo{person}{Minseop Park}, \bibinfo{person}{Jaeseong You}, \bibinfo{person}{Markus Nagel}, {and} \bibinfo{person}{Simyung Chang}.} \bibinfo{year}{2022}\natexlab{}.
\newblock \showarticletitle{Quadapter: Adapter for gpt-2 quantization}.
\newblock \bibinfo{journal}{\emph{arXiv preprint arXiv:2211.16912}} (\bibinfo{year}{2022}).
\newblock


\bibitem[Pignatelli et~al\mbox{.}(2024)]%
        {pignatelli2024assessing}
\bibfield{author}{\bibinfo{person}{Eduardo Pignatelli}, \bibinfo{person}{Johan Ferret}, \bibinfo{person}{Tim Rock{\"a}schel}, \bibinfo{person}{Edward Grefenstette}, \bibinfo{person}{Davide Paglieri}, \bibinfo{person}{Samuel Coward}, {and} \bibinfo{person}{Laura Toni}.} \bibinfo{year}{2024}\natexlab{}.
\newblock \showarticletitle{Assessing the zero-shot capabilities of LLMs for action evaluation in RL}.
\newblock \bibinfo{journal}{\emph{arXiv preprint arXiv:2409.12798}} (\bibinfo{year}{2024}).
\newblock


\bibitem[{PyTorch Core Team}(2024)]%
        {pytorchhub}
\bibfield{author}{\bibinfo{person}{{PyTorch Core Team}}.} \bibinfo{year}{2024}\natexlab{}.
\newblock \bibinfo{title}{PyTorch Hub}.
\newblock \bibinfo{howpublished}{\url{https://pytorch.org/hub}}.
\newblock


\bibitem[Riva et~al\mbox{.}(2024)]%
        {riva2024huggingface}
\bibfield{author}{\bibinfo{person}{Matteo Riva}, \bibinfo{person}{Tommaso~Lorenzo Parigi}, \bibinfo{person}{Federica Ungaro}, {and} \bibinfo{person}{Luca Massimino}.} \bibinfo{year}{2024}\natexlab{}.
\newblock \showarticletitle{HuggingFace's impact on medical applications of artificial intelligence}.
\newblock \bibinfo{journal}{\emph{Computational and Structural Biotechnology Reports}} (\bibinfo{year}{2024}), \bibinfo{pages}{100003}.
\newblock


\bibitem[R{\"o}ttger et~al\mbox{.}(2024)]%
        {rottger2024safetyprompts}
\bibfield{author}{\bibinfo{person}{Paul R{\"o}ttger}, \bibinfo{person}{Fabio Pernisi}, \bibinfo{person}{Bertie Vidgen}, {and} \bibinfo{person}{Dirk Hovy}.} \bibinfo{year}{2024}\natexlab{}.
\newblock \showarticletitle{Safetyprompts: a systematic review of open datasets for evaluating and improving large language model safety}.
\newblock \bibinfo{journal}{\emph{arXiv preprint arXiv:2404.05399}} (\bibinfo{year}{2024}).
\newblock


\bibitem[Shen et~al\mbox{.}(2023)]%
        {shen2023slimpajama}
\bibfield{author}{\bibinfo{person}{Zhiqiang Shen}, \bibinfo{person}{Tianhua Tao}, \bibinfo{person}{Liqun Ma}, \bibinfo{person}{Willie Neiswanger}, \bibinfo{person}{Zhengzhong Liu}, \bibinfo{person}{Hongyi Wang}, \bibinfo{person}{Bowen Tan}, \bibinfo{person}{Joel Hestness}, \bibinfo{person}{Natalia Vassilieva}, \bibinfo{person}{Daria Soboleva}, {et~al\mbox{.}}} \bibinfo{year}{2023}\natexlab{}.
\newblock \showarticletitle{Slimpajama-dc: Understanding data combinations for llm training}.
\newblock \bibinfo{journal}{\emph{arXiv preprint arXiv:2309.10818}} (\bibinfo{year}{2023}).
\newblock


\bibitem[Singla et~al\mbox{.}(2023)]%
        {singla2023empirical}
\bibfield{author}{\bibinfo{person}{Tanmay Singla}, \bibinfo{person}{Dharun Anandayuvaraj}, \bibinfo{person}{Kelechi~G Kalu}, \bibinfo{person}{Taylor~R Schorlemmer}, {and} \bibinfo{person}{James~C Davis}.} \bibinfo{year}{2023}\natexlab{}.
\newblock \showarticletitle{An empirical study on using large language models to analyze software supply chain security failures}. In \bibinfo{booktitle}{\emph{Proceedings of the 2023 Workshop on Software Supply Chain Offensive Research and Ecosystem Defenses}}. \bibinfo{pages}{5--15}.
\newblock


\bibitem[Sonatype(2015)]%
        {sonatype2015software}
\bibfield{author}{\bibinfo{person}{Sonatype}.} \bibinfo{year}{2015}\natexlab{}.
\newblock \bibinfo{booktitle}{\emph{2015 State of the Software Supply Chain Report}}.
\newblock \bibinfo{type}{{T}echnical {R}eport}. \bibinfo{institution}{Sonatype}.
\newblock
\urldef\tempurl%
\url{https://www.sonatype.com/hubfs/White_Papers/2015_State_of_the_Software_Supply_Chain_Report-.pdf}
\showURL{%
\tempurl}


\bibitem[Tan et~al\mbox{.}(2024)]%
        {tan2024challenges}
\bibfield{author}{\bibinfo{person}{Xin Tan}, \bibinfo{person}{Taichuan Li}, \bibinfo{person}{Ruohe Chen}, \bibinfo{person}{Fang Liu}, {and} \bibinfo{person}{Li Zhang}.} \bibinfo{year}{2024}\natexlab{}.
\newblock \showarticletitle{Challenges of Using Pre-trained Models: the Practitioners' Perspective}.
\newblock \bibinfo{journal}{\emph{arXiv preprint arXiv:2404.14710}} (\bibinfo{year}{2024}).
\newblock


\bibitem[Thakkar and Manimaran(2023)]%
        {thakkar2023comprehensive}
\bibfield{author}{\bibinfo{person}{Hiren Thakkar} {and} \bibinfo{person}{A Manimaran}.} \bibinfo{year}{2023}\natexlab{}.
\newblock \showarticletitle{Comprehensive examination of instruction-based language models: A comparative analysis of mistral-7b and llama-2-7b}. In \bibinfo{booktitle}{\emph{2023 International Conference on Emerging Research in Computational Science (ICERCS)}}. IEEE, \bibinfo{pages}{1--6}.
\newblock


\bibitem[Vasiliev(2020)]%
        {vasiliev2020natural}
\bibfield{author}{\bibinfo{person}{Yuli Vasiliev}.} \bibinfo{year}{2020}\natexlab{}.
\newblock \bibinfo{booktitle}{\emph{Natural language processing with Python and spaCy: A practical introduction}}.
\newblock \bibinfo{publisher}{No Starch Press}.
\newblock


\bibitem[Velingker et~al\mbox{.}({[n.\,d.]})]%
        {velingkerclam}
\bibfield{author}{\bibinfo{person}{Neelay Velingker}, \bibinfo{person}{Jason Liu}, \bibinfo{person}{Amish Sethi}, \bibinfo{person}{William Dodds}, \bibinfo{person}{Zhiqiu Xu}, \bibinfo{person}{Saikat Dutta}, \bibinfo{person}{Mayur Naik}, {and} \bibinfo{person}{Eric Wong}.} \bibinfo{year}{[n.\,d.]}\natexlab{}.
\newblock \showarticletitle{CLAM: Unifying Finetuning, Quantization, and Pruning by Chaining LLM Adapter Modules}. In \bibinfo{booktitle}{\emph{Workshop on Efficient Systems for Foundation Models II@ ICML2024}}.
\newblock


\bibitem[Villalobos et~al\mbox{.}(2024)]%
        {villalobos2024will}
\bibfield{author}{\bibinfo{person}{Pablo Villalobos}, \bibinfo{person}{Anson Ho}, \bibinfo{person}{Jaime Sevilla}, \bibinfo{person}{Tamay Besiroglu}, \bibinfo{person}{Lennart Heim}, {and} \bibinfo{person}{Marius Hobbhahn}.} \bibinfo{year}{2024}\natexlab{}.
\newblock \showarticletitle{Will we run out of data? Limits of LLM scaling based on human-generated data}.
\newblock \bibinfo{journal}{\emph{arXiv preprint arXiv:2211.04325}} (\bibinfo{year}{2024}), \bibinfo{pages}{13--29}.
\newblock


\bibitem[VM et~al\mbox{.}(2024)]%
        {vm2024fine}
\bibfield{author}{\bibinfo{person}{Kushala VM}, \bibinfo{person}{Harikrishna Warrier}, \bibinfo{person}{Yogesh Gupta}, {et~al\mbox{.}}} \bibinfo{year}{2024}\natexlab{}.
\newblock \showarticletitle{Fine Tuning LLM for Enterprise: Practical Guidelines and Recommendations}.
\newblock \bibinfo{journal}{\emph{arXiv preprint arXiv:2404.10779}} (\bibinfo{year}{2024}).
\newblock


\bibitem[Wang et~al\mbox{.}(2025)]%
        {wang2025bingo}
\bibfield{author}{\bibinfo{person}{Pinhuan Wang}, \bibinfo{person}{Chengying Huan}, \bibinfo{person}{Zhibin Wang}, \bibinfo{person}{Chen Tian}, \bibinfo{person}{Yuede Ji}, {and} \bibinfo{person}{Hang Liu}.} \bibinfo{year}{2025}\natexlab{}.
\newblock \showarticletitle{Bingo: Radix-based Bias Factorization for Random Walk on Dynamic Graphs}. In \bibinfo{booktitle}{\emph{Proceedings of the Twentieth European Conference on Computer Systems}} (Rotterdam, Netherlands) \emph{(\bibinfo{series}{EuroSys '25})}. \bibinfo{publisher}{Association for Computing Machinery}, \bibinfo{numpages}{16}~pages.
\newblock
\showISBNx{9798400711961}


\bibitem[Wang et~al\mbox{.}(2024)]%
        {wang2024large}
\bibfield{author}{\bibinfo{person}{Shenao Wang}, \bibinfo{person}{Yanjie Zhao}, \bibinfo{person}{Xinyi Hou}, {and} \bibinfo{person}{Haoyu Wang}.} \bibinfo{year}{2024}\natexlab{}.
\newblock \showarticletitle{Large language model supply chain: A research agenda}.
\newblock \bibinfo{journal}{\emph{ACM Transactions on Software Engineering and Methodology}} (\bibinfo{year}{2024}).
\newblock


\bibitem[Xu et~al\mbox{.}(2024b)]%
        {xu2024survey}
\bibfield{author}{\bibinfo{person}{Mengwei Xu}, \bibinfo{person}{Wangsong Yin}, \bibinfo{person}{Dongqi Cai}, \bibinfo{person}{Rongjie Yi}, \bibinfo{person}{Daliang Xu}, \bibinfo{person}{Qipeng Wang}, \bibinfo{person}{Bingyang Wu}, \bibinfo{person}{Yihao Zhao}, \bibinfo{person}{Chen Yang}, \bibinfo{person}{Shihe Wang}, {et~al\mbox{.}}} \bibinfo{year}{2024}\natexlab{b}.
\newblock \showarticletitle{A survey of resource-efficient llm and multimodal foundation models}.
\newblock \bibinfo{journal}{\emph{arXiv preprint arXiv:2401.08092}} (\bibinfo{year}{2024}).
\newblock


\bibitem[Xu et~al\mbox{.}(2024a)]%
        {xu2024towards}
\bibfield{author}{\bibinfo{person}{Zhuoyan Xu}, \bibinfo{person}{Zhenmei Shi}, \bibinfo{person}{Junyi Wei}, \bibinfo{person}{Fangzhou Mu}, \bibinfo{person}{Yin Li}, {and} \bibinfo{person}{Yingyu Liang}.} \bibinfo{year}{2024}\natexlab{a}.
\newblock \showarticletitle{Towards Few-Shot Adaptation of Foundation Models via Multitask Finetuning}.
\newblock \bibinfo{journal}{\emph{arXiv preprint arXiv:2402.15017}} (\bibinfo{year}{2024}).
\newblock


\bibitem[Yenduri et~al\mbox{.}(2024)]%
        {yenduri2024gpt}
\bibfield{author}{\bibinfo{person}{Gokul Yenduri}, \bibinfo{person}{M Ramalingam}, \bibinfo{person}{G~Chemmalar Selvi}, \bibinfo{person}{Y Supriya}, \bibinfo{person}{Gautam Srivastava}, \bibinfo{person}{Praveen Kumar~Reddy Maddikunta}, \bibinfo{person}{G~Deepti Raj}, \bibinfo{person}{Rutvij~H Jhaveri}, \bibinfo{person}{B Prabadevi}, \bibinfo{person}{Weizheng Wang}, {et~al\mbox{.}}} \bibinfo{year}{2024}\natexlab{}.
\newblock \showarticletitle{Gpt (generative pre-trained transformer)--a comprehensive review on enabling technologies, potential applications, emerging challenges, and future directions}.
\newblock \bibinfo{journal}{\emph{IEEE Access}} (\bibinfo{year}{2024}).
\newblock


\bibitem[Zou et~al\mbox{.}(2023)]%
        {zou2023comprehensive}
\bibfield{author}{\bibinfo{person}{Wentao Zou}, \bibinfo{person}{Qi Li}, \bibinfo{person}{Jidong Ge}, \bibinfo{person}{Chuanyi Li}, \bibinfo{person}{Xiaoyu Shen}, \bibinfo{person}{Liguo Huang}, {and} \bibinfo{person}{Bin Luo}.} \bibinfo{year}{2023}\natexlab{}.
\newblock \showarticletitle{A Comprehensive Evaluation of Parameter-Efficient Fine-Tuning on Software Engineering Tasks}.
\newblock \bibinfo{journal}{\emph{arXiv preprint arXiv:2312.15614}} (\bibinfo{year}{2023}).
\newblock


\bibitem[Čolaković et~al\mbox{.}(2021)]%
        {colakovic2021traditional}
\bibfield{author}{\bibinfo{person}{Ajdin Čolaković}, \bibinfo{person}{Aleksandar Đorđević}, \bibinfo{person}{Branislav Cvetić}, \bibinfo{person}{Milan Danilović}, {and} \bibinfo{person}{Dejan Vasiljević}.} \bibinfo{year}{2021}\natexlab{}.
\newblock \showarticletitle{Traditional vs Digital Supply Chains}.
\newblock \bibinfo{journal}{\emph{ResearchGate}} (\bibinfo{year}{2021}).
\newblock
\urldef\tempurl%
\url{https://www.researchgate.net/publication/381617842_Traditional_vs_Digital_Supply_Chains}
\showURL{%
\tempurl}


\end{thebibliography}

\end{document}